\documentclass[11pt]{article}

\usepackage[final]{acl}
\usepackage{authblk}

\usepackage{times}
\usepackage{latexsym}
\usepackage{amsfonts}
\usepackage{booktabs}
\usepackage{amsmath}
\usepackage{colortbl} % 用于给"Ours"那一行上色
\usepackage{multirow}
\usepackage[table]{xcolor}

\usepackage[T1]{fontenc}
\usepackage[utf8]{inputenc}

\usepackage{microtype}
\usepackage{tikz}
\usetikzlibrary{positioning,arrows.meta,fit}

\usepackage{inconsolata}

\usepackage{graphicx}

\title{QA-MoE: Towards a Continuous Reliability Spectrum with Quality-Aware Mixture of Experts for Robust Multimodal Sentiment Analysis}

\author[1]{Yitong Zhu}
\author[2]{Yuxuan Jiang}
\author[1]{Guanxuan Jiang}
\author[1]{Bojing Hou}
\author[3]{Peng Yuan Zhou}
\author[1]{Ge Lin KAN}
\author[1, *]{Yuyang Wang}
\affil[1]{The Hong Kong University of Science and Technology (Guangzhou)}
\affil[2]{Tsinghua University}
\affil[3]{Aarhus University}

\affil[ ]{\texttt{\{yzhu162, gjiang240, bhou870\}@connect.hkust-gz.edu.cn}}
\affil[]{\texttt{\{gelin,yuyangwang\}@hkust-gz.edu.cn}}
\affil[ ]{\texttt{jiangyux25@mails.tsinghua.edu.cn}, \texttt{pyzhou@cs.au.dk}}

\date{} % 如果不需要显示日期，请加上这一行

\begin{document}
\maketitle
\begin{abstract}
 
Multimodal Sentiment Analysis (MSA) aims to infer human sentiment from textual, acoustic, and visual signals. In real-world scenarios, however, multimodal inputs are often compromised by dynamic noise or modality missingness. Existing methods typically treat these imperfections as discrete cases or assume fixed corruption ratios, which limits their adaptability to continuously varying reliability conditions.
To address this, we first introduce a Continuous Reliability Spectrum to unify missingness and quality degradation into a single framework. Building on this, we propose QA-MoE, a Quality-Aware Mixture-of-Experts framework that quantifies modality reliability via self-supervised aleatoric uncertainty. This mechanism explicitly guides expert routing, enabling the model to suppress error propagation from unreliable signals while preserving task-relevant information.
Extensive experiments indicate that QA-MoE achieves competitive or state-of-the-art performance across diverse degradation scenarios and exhibits a promising One-Checkpoint-for-All property in practice.
\end{abstract}

\section{Introduction}
\label{sec:intro}
Multimodal Sentiment Analysis (MSA) is a base of human-centric computing, which aims to explain complex emotional states by integrating Textual ($T$), Vision ($V$), and Acoustic ($A$)~\cite{Sun2025sequential, Yang2024clgsi, zhang2025qammslf, He2025MSAmba, jiang2026trust}. With recent advances in multimodal learning~\cite{Chen2023InternVS, Mizrahi20234M, zhu2024languagebind}, MSA models ~\cite{Fang2025EMOE, Kang2025PaSE} can better exploit the complementary signals across modalities and capture subtle affective cues that unimodal systems often miss, narrowing the gap between human expression and machine understanding.

\begin{figure}[t]
    \centering
    \includegraphics[width=\linewidth]{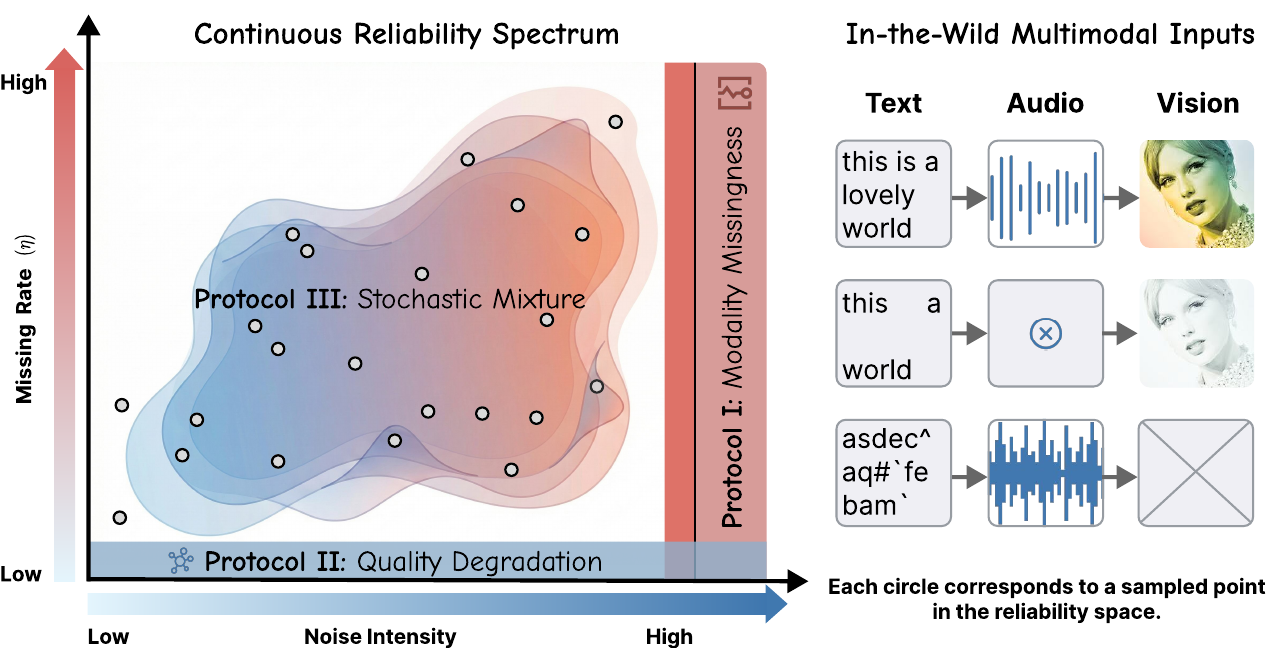} 
    \caption{The Continuous Reliability Spectrum unifies three evaluation protocols defined by noise intensity ($\lambda$) and missing rate ($\eta$), and inputs from Text, Audio, and Vision are processed as imperfect multimodal inputs.}
    \label{fig:spectrum}
\end{figure}

However, unlike the clean and complete data found in laboratory settings, real-world multimodal signals are often noisy or incomplete. Most existing models rely on the assumption of ideal inputs, creating a significant gap between constrained training conditions and the complexities of practical application~\cite{Zhao2021MMIN, Xu2024MoMKE}. In practice, modality noise fluctuates dynamically due to environmental interference, while data incompleteness frequently arises from sensor failures. These uncertainties often manifest as a stochastic mixture, where noise and missingness co-occur non-uniformly across samples (Figure.~\ref{fig:spectrum}). Crucially, these data defects are not discrete categories, but instead exist across a broad range of intensities, ranging from subtle noise to the total loss of signal.

To tackle these reliability issues, earlier efforts have primarily focused on explicit data imputation, employing reconstruction methods~\cite {Cai2018DALMMD, Lian2023GCNet, Guo2024MPLMM} to recover missing modalities from the remaining observed signals. Recent advancements have pivoted toward architectural robustness, leveraging Bayesian meta-learning~\cite{Ma2021SMILML}, diffusion models~\cite{Wang2024IMDER}, and attention mechanisms~\cite{mai2025supervised} to directly learn from imperfect inputs. Nevertheless, these approaches still suffer from two critical limitations: (1) \textit{Quality-Agnostic Extraction.} Existing models typically derive representations from raw inputs without explicitly modeling their reliability. Consequently, they fail to disentangle task-relevant semantics from non-informative noise, causing the model to capture spurious noisy artifacts rather than robust affective cues. (2) \textit{Fixed-Ratio Bias.} These methods are optimized for specific, predefined corruption ratios seen during training. This rigidity prevents them from adapting to the fluctuating noise intensities in the wild, leading to severe performance drops when test-time reliability deviates from training protocols.

To bridge these gaps, we propose a unified framework centered around a \textbf{Continuous Reliability Spectrum} (Figure.~\ref{fig:spectrum}). Rather than treating data imperfections as discrete cases, we conceptually map diverse defects onto this continuous spectrum, unifying three distinct evaluation protocols: \textit{Modality Missingness}, \textit{Quality Degradation}, and \textit{Stochastic Mixture}. This perspective allows us to evaluate model robustness in a more holistic and realistic manner.
Building upon this unified view, we introduce the \textbf{Quality-Aware Mixture of Experts} (\textit{QA-MoE}). To move beyond traditional semantic-only routing, we incorporate a self-supervised reliability quantification module that utilizes aleatoric uncertainty to generate dynamic quality scores. These scores explicitly guide the MoE computation, transforming the routing process into a quality-aware aggregation. By weighting semantic gating with these reliability metrics, the framework effectively suppresses expert activation for unreliable inputs while prioritizing task-relevant signals. 
Finally, extensive experiments on MSA and Multimodal Intent Recognition (MIR) tasks demonstrate that QA-MoE achieves superior performance, validating the robustness and versatility of the proposed framework. Notably, evaluations across the comprehensive settings of the reliability spectrum reveal that our model effectively navigates varying levels of noise and missingness, establishing a \textbf{One-Checkpoint-for-All} capability. This signifies that a single trained model can generalize to arbitrary, unseen degradation intensities without retraining or specialized fine-tuning.
The main contributions of our work are summarized as follows:

\begin{itemize}
\item We propose the Continuous Reliability Spectrum to unify modality missingness and quality degradation into a framework, moving beyond traditional treatment of discrete defects.
% \item We propose QA-MoE, a quality-aware mixture-of-experts framework in which self-supervised reliability estimation, derived from aleatoric uncertainty, explicitly guides routing to adapt expert aggregation to input quality.
\item We introduce a quality-aware mixture-of-experts framework, QA-MoE, in which self-supervised reliability estimation derived from aleatoric uncertainty explicitly guides routing to adapt expert aggregation to input quality.
\item Extensive experiments on MSA benchmarks (CMU-MOSI, CMU-MOSEI) and cross-task datasets (IEMOCAP, MIntRec) show that our model achieves state-of-the-art performance, and exhibits a One-Checkpoint-for-All capability, generalizing across a wide range of noise levels and missing rates.
\end{itemize}

\section{Related Work}
\subsection{Multimodal Sentiment Analysis}

MSA integrates heterogeneous signals from language, vision, and acoustics to infer human emotions. Early work modeled explicit cross-modal interactions via tensor fusion~\cite{Zadeh2017TensorFN}. Transformer-based architectures~\cite{Tsai2019MulT, zhang2023metatransformer} later advanced the field through cross-modal attention for aligning asynchronous streams. More recently, representation learning approaches~\cite{zhu2025MISRL} have emphasized disentanglement~\cite{Hazarika2020MISA, zhu2025hierarchicalmoecontinuousmultimodal} and self-supervised objectives~\cite{Yang2024clgsi, He2025MSAmba} to reduce redundancy. However, most methods still assume complete and noise-free modalities. In contrast, we develop a unified framework that explicitly estimates signal reliability and operates across a continuous spectrum of degradation.

\subsection{Imperfect Multimodal Learning}

Real-world deployments often involve missing or noisy modalities. Early work addressed these imperfections through data imputation~\cite{Cai2018DALMMD, Ma2021SMILML}, reconstructing missing views via generative models; however, such methods are computationally expensive and vulnerable to distribution shift–induced hallucination~\cite{Guo2024MPLMM}. Subsequent research shifted toward robust representation learning, down-weighting corrupted features via attention~\cite{mai2025supervised} or auxiliary objectives~\cite{Wang2024IMDER}, yet these approaches still struggle under severe noise. More recently, Mixture-of-Experts or Adapters have been adopted for multimodal learning~\cite{Xu2024MoMKE, Chen2025MMA}, but existing routers are purely semantics-driven and fail to distinguish informative signals from corruption. To address this, we introduce a self-supervised quality signal into the routing process in order to enable effective isolation of unreliable modalities.

\begin{figure*}[t]
    \centering
    \includegraphics[width=1.00\linewidth]{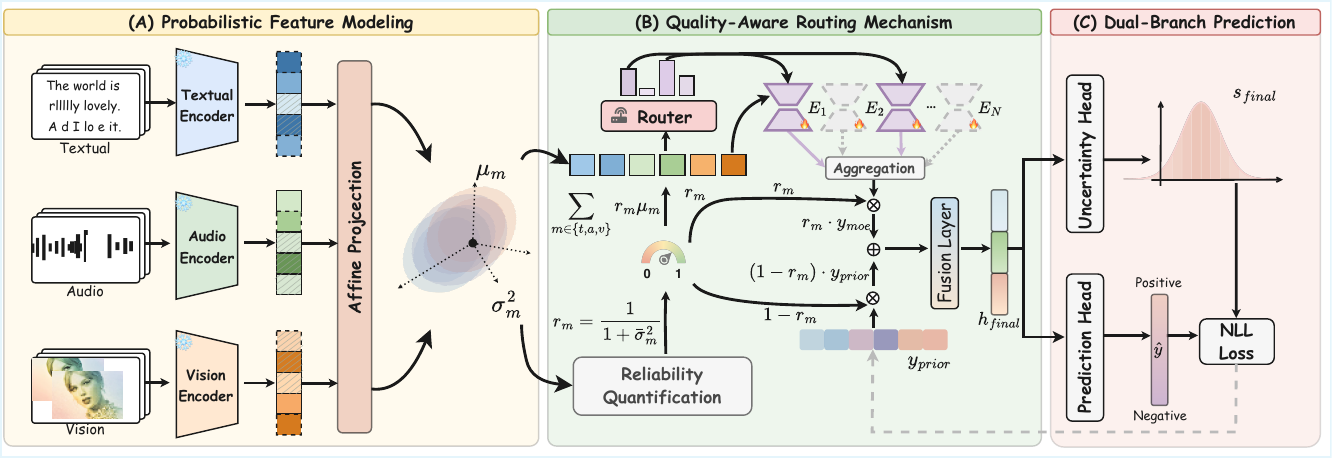} % 替换你的文件名
    \caption{Overview of QA-MoE.
(A) Probabilistic Feature Modeling encodes inputs as distributions to capture the uncertainty.
(B) Quality-Aware Routing calculates a quality score $r_m$ to guide the expert selection.
(C) Dual-Branch Prediction outputs both the prediction and uncertainty to optimize the model via heteroscedastic regression.}
    \label{fig:main_arch}
\end{figure*}

\section{Methodology}
\label{sec:methodology}

\subsection{Preliminaries}
\label{sec:preliminaries}

To bridge the gap between idealized laboratory benchmarks and unpredictable real-world scenarios, we establish a foundational framework from standard encoding to  reliability analysis.

\noindent \textbf{Standard Multimodal Encoding.}
Given a multimodal dataset $\mathcal{D} = \{(X_i, y_i)\}_{i=1}^N$, each sample $X_i$ comprises Textual ($t$), Audio ($a$), and Video ($v$) modalities. Following standard protocols~\cite{Tsai2019MulT}, we employ pre-trained unimodal encoders $E_{m \in \{t, a, v\}}$ to extract the raw feature sequences $\mathbf{u}_m \in \mathbb{R}^{T_m \times d_m}$. In conventional lab-controlled settings, these extracted features are implicitly assumed to be complete and pristine.

\noindent \textbf{From Ideal to Real.}
However, to capture the dynamic nature of real-world noise, we depart from this idealization. We first introduce the \textit{Continuous Reliability Spectrum} (Sec.~\ref{sec:spectrum_theory}) to conceptually unify diverse data defects. Subsequently, we formalize this concept through \textit{Stochastic Imperfection Modeling} (Sec.~\ref{sec:stochastic_modeling}), which provides the theoretical basis for our evaluation protocols.

% In order to bridge the gap between idealized laboratory benchmarks and unpredictable real-world deployments. We first define the standard encoding process and then define the Stochastic Imperfection Modeling, which provides the theoretical basis for our evaluation protocols.

% Given a multimodal dataset $\mathcal{D} = \{X_i, Y_i\}_{i=1}^N$, each sample consists of Textual ($T$), Vision ($V$), and Acoustic ($A$) modalities. 
% Same with~\cite{Tsai2019MulT}, we employ pre-trained encoders $E_{m\in D}(\cdot)$ to extract unimodal feature vectors $\mathbf{u}_m \in \mathbb{R}^{d_m}$. In lab-controlled settings, they are considered complete and clean.

\subsection{Continuous Reliability Spectrum}
\label{sec:spectrum_theory}
Considering that noise and missingness often occur simultaneously in real-world scenarios, we propose a unified reliability spectrum shown in Figure~\ref{fig:spectrum}(left) to conceptually map these diverse imperfections onto a single latent measure. 
Instead of treating defects as discrete categories, we quantify the input quality by a \textbf{Latent Reliability Score} $r_m \in (0, 1]$:
\begin{equation}
    \text{Degradation} \propto 1 - r_m
\end{equation}
We define three characteristic phases along this spectrum based on $r_{m}$:
\textbf{High Quality ($r_m \approx 1$)}: Ideal clean data found in lab settings.
\textbf{Quality Degradation ($r_m \in (0,1)$)}: Data corrupted by noise intensity $\lambda_m$.
\textbf{Availability Limit ($r_m \to 0$)}: Data subject to modality missingness rate $\eta$.

\subsection{Stochastic Imperfection Modeling}\label{sec:stochastic_modeling}Based on the unified spectrum, we formulate the real-world environment not as a static dataset, but as a \textbf{Stochastic Degradation Process}. For any input sample $\mathbf{u}_m$, the imperfect representation $\tilde{\mathbf{u}}_m$ is generated via a transformation function $\mathcal{T}$:
\begin{equation}
\tilde{\mathbf{u}}_m = (1 - \mathbb{I}_{\mathrm{miss}}) \cdot (\mathbf{u}_m + \boldsymbol{\epsilon}_m)
\label{eq:transformation}
\end{equation}
$\mathbb{I}_{\mathrm{miss}} \sim \operatorname{Bernoulli}(\eta)$ here is a binary variable indicating modality absence.The term $\boldsymbol{\epsilon}_m \sim \mathcal{N}(\mathbf{0}, \sigma^2(\lambda_m)\mathbf{I})$ represents the additive noise, where the variance is governed by the degradation intensity $\lambda_m$.This formulation unifies three distinct evaluation protocols representing different subspaces of the reliability spectrum:

\noindent \textbf{Protocol I: Modality Missingness.} We focus on binary availability by setting $\boldsymbol{\epsilon}_m = \mathbf{0}$ and varying the missing rate $\eta$ to simulate scenarios such as sensor failure or packet loss.

\noindent \textbf{Protocol II: Quality Degradation.} We focus on signal fidelity by fixing $\mathbb{I}_{\mathrm{miss}} = 0$ and varying the noise intensity $\lambda_m$. This simulates noisy environments where the modality is present but unreliable.

\noindent \textbf{Protocol III: Stochastic Mixture.} We sample both $\lambda_m$ and $\eta$ from a joint distribution $P_{env}(\lambda, \eta)$. In this setting, input signals are subject to random combinations of noise corruption and modality unavailability, reflecting complex in-the-wild dynamics.

\subsection{Quality-Aware Mixture of Experts}
\label{sec:qa_moe_framework}

To effectively navigate the continuous reliability spectrum, we introduce the QA-MoE framework. 
Unlike standard deterministic models~\cite{MMKT2025Shi, MSA2024Zhang}, QA-MoE operates under a probabilistic principle: it decouples the input representation into a semantic signal and an uncertainty measure, which is used to explicitly guide the computation flow.

\subsubsection{Probabilistic Feature Modeling}
\label{sec:probabilistic_modeling}
Under our stochastic degradation process (Sec. \ref{sec:stochastic_modeling}), $\mathbf{u}_m$ is inevitably influenced by varying degrees of noise. To model the inherent uncertainty, we project the feature space onto a multivariate Gaussian distribution:
\begin{equation}
    p(\mathbf{z}_m | \mathbf{u}_m) = \mathcal{N}(\mathbf{z}_m; \boldsymbol{\mu}_m, \text{diag}(\boldsymbol{\sigma}^2_m))
\end{equation}
, where $\mathbf{z}_m$ is the latent representation. 
We employ two parallel affine transformations to estimate the distribution parameters:
\begin{align}
    \boldsymbol{\mu}_m &= \mathbf{W}_\mu \mathbf{u}_m + \mathbf{b}_\mu \\
    \boldsymbol{\sigma}^2_m &= \text{Softplus}(\mathbf{W}_\sigma \mathbf{u}_m + \mathbf{b}_\sigma)
\end{align}
, $\mathbf{W}_{\mu/\sigma} \in \mathbb{R}^{d \times d}$ and $\mathbf{b}_{\mu/\sigma} \in \mathbb{R}^{d}$ here are learnable projection matrices and bias terms. Specifically, $\boldsymbol{\mu}_m$ extracts the clean semantic signal, while $\boldsymbol{\sigma}^2_m$ quantifies the inherent uncertainty. High variance values indicate unreliable or missing inputs, serving as a dynamic quality indicator to guide the subsequent routing.

\subsubsection{Quality-Aware Routing Mechanism}
\label{sec:routing_mechanism}
To solve the limitations of static computation, we design a quality-aware adaptive routing mechanism to dynamically align expert contribution with input reliability. It involves two key steps: quality quantification and modulating the expert aggregation.

\noindent \textbf{Quality Quantification.}
First, we derive a scalar quality score $r_m \in (0, 1]$ to explicitly quantify the reliability of the input modality. 
Since the variance $\boldsymbol{\sigma}^2_m$ serves as an indicator of uncertainty, the quality is naturally modeled as inversely proportional to the aggregate variance:
\begin{equation}
    r_m = \frac{1}{1 + \frac{1}{d} \sum_{k=1}^{d} \boldsymbol{\sigma}^2_{m,k}}
\end{equation}
This formulation creates a bounded metric: when the input is clean, $r_m \to 1$; conversely, as degradation intensifies and variance explodes, $r_m$ asymptotically decays to 0.

\noindent \textbf{Selective Expert Aggregation.}
Next, we integrate the score into the MoE computation. We employ a bank of $N$ experts $\{E_i\}_{i=1}^N$, where each expert is instantiated as a GLU to capture complex semantic patterns.
A routing network computes the gating weights $g(\boldsymbol{\mu}_m) = \text{Softmax}(\mathbf{W}_g \boldsymbol{\mu}_m)$ based on the semantic centroid.

Crucially, we introduce $r_m$ as a global suppression coefficient and the output $\mathbf{y}_m$ is computed via a smooth interpolation:
\begin{equation}
    \mathbf{y}_m = r_m \cdot \sum_{i=1}^N g_i(\boldsymbol{\mu}_m) E_i(\boldsymbol{\mu}_m) + (1 - r_m) \cdot \mathbf{y}_{prior}
\end{equation}
Here, $\mathbf{y}_{prior}$ is a learnable global static embedding, which captures a dataset-level semantic consensus independent of specific inputs. This allows the model to smoothly interpolate between the instance-specific prediction and this stable reference to mitigate the risk of overfitting to noise.

\subsubsection{Dual-Branch Prediction}
\label{sec:final_prediction}

After obtaining the refined expert outputs $\mathbf{y}_m$ for all modalities, we aggregate them into a unified multimodal representation $\mathbf{h}_{final}$ via a standard fusion layer.

To enable the heteroscedastic regression objective (detailed in Sec. \ref{sec:optimization}), the model must output not only a sentiment score but also a confidence measure. 
Therefore, we design a dual-branch regression head that decouples the prediction of value and total uncertainty:
\begin{align}
    \hat{y} &= \mathbf{W}_y \mathbf{h}_{final} + b_y \\
    \mathbf{s}_{final} &= \mathbf{W}_s \mathbf{h}_{final} + b_s
\end{align}
Here, $\hat{y}$ and $\mathbf{s}_{final}$ is the predicted sentiment score and log-variance. $\mathbf{s}_{final}$ acts as a learned estimator of the total prediction uncertainty for the current sample, which is used to dynamically weigh the gradient updates during optimization.

\subsection{Training and Optimization}
\label{sec:optimization}
To empower QA-MoE with the capability to generalize across the continuous reliability spectrum in Sec.~\ref{sec:intro}, we propose a unified learning framework that integrates a dynamic data augmentation strategy with a uncertainty-aware objective function.

\noindent \textbf{Spectrum-Aware Training Strategy.}
Instead of relying on perfect datasets, we construct a dynamic training process to simulate real-world imperfections according to Sec~\ref{sec:stochastic_modeling}. Specifically, for each training batch, we randomly inject noise and mask modalities with varying probabilities. This exposes the router to the full reliability spectrum during optimization. Detailed protocols for Spectrum Dataset generation are provided in Appendix~\ref{sec:appendix_spectrum_generation}.

\noindent \textbf{Optimization Objectives.}
To effectively train the QA-MoE framework under the Stochastic Degradation Protocol, we treat the model's output as a Gaussian distribution rather than a deterministic point estimate.
$\hat{y}$ and $\mathbf{s}_{final}$ derived in Sec. \ref{sec:final_prediction} are used to minimize the Negative Log-Likelihood (NLL) of the ground truth $y$.
The total loss $\mathcal{L}$ establishes a self-supervised feedback loop, which is formulated as:
\begin{equation}
    \mathcal{L} = \frac{1}{N} \sum_{i=1}^{N} \left( \frac{1}{2} e^{-\mathbf{s}_{final, i}} (y_i - \hat{y}_i)^2 + \frac{1}{2} \mathbf{s}_{final, i} \right)
\end{equation}

For noisy inputs with large errors, reducing $\mathcal{L}$ forces $\mathbf{s}_{final}$ to increase. The gradient backpropagates to elevate the variance $\boldsymbol{\sigma}^2$, which directly reduces the quality score $r_{m}$. 
Consequently, the router learns to suppress experts for degraded data without manual labels.

\section{Experiments}

\begin{table*}[t]
\centering
\renewcommand{\arraystretch}{0.75}
% 建议将宽度调整为 1.0\textwidth 以容纳更多列，避免字体过小
\resizebox{1.0\textwidth}{!}{
\begin{tabular}{l|ccccc|ccccc|cc}
\toprule
\multirow{2}{*}{\textbf{Models}} & \multicolumn{5}{c|}{\textbf{CMU-MOSI}} & \multicolumn{5}{c|}{\textbf{CMU-MOSEI}} & \multicolumn{2}{c}{\textbf{MIntRec}} \\
 & $\text{ACC}_{7}$ $\uparrow$ & $\text{ACC}_{2}$ $\uparrow$ & $\text{F}_{1}$ $\uparrow$ & MAE $\downarrow$ & Corr $\uparrow$ & $\text{ACC}_{7}$ $\uparrow$ & $\text{ACC}_{2}$ $\uparrow$ & $\text{F}_{1}$ $\uparrow$ & MAE $\downarrow$ & Corr $\uparrow$ & ACC $\uparrow$ & $\text{F}_{1}$ $\uparrow$ \\
\midrule
TFN$^\dagger$   & 31.9 & 78.8 & 78.9 & 0.953 & 0.698 & 50.9 & 80.4 & 80.7 & 0.574 & 0.700 & - & - \\
LMF$^\dagger$   & 36.9 & 78.7 & 78.7 & 0.931 & 0.695 & 52.3 & 84.7 & 84.5 & 0.564 & 0.677 & - & - \\
MulT$^\dagger$  & 35.1 & 80.0 & 80.1 & 0.936 & 0.711 & 52.3 & 82.7 & 82.8 & 0.572 &   -   & 72.6 & 69.5 \\
MISA$^\dagger$  & 41.8 & 84.2 & 84.2 & 0.754 & 0.761 & 52.3 & 85.3 & 85.1 & 0.543 & 0.756 & 72.4 & 70.8 \\
MMIM$^\dagger$  & 45.8 & 84.6 & 84.5 & 0.717 &   -   & 50.1 & 83.6 & 83.5 & 0.580 &   -   & - & - \\
EMOE$^\dagger$  & 47.7 & 85.4 & 85.4 & 0.710 &   -   & 54.1 & 85.3 & 85.3 & 0.536 &   -   & 72.6 & 70.7 \\
MMA$^\ddagger$   & 46.9 & 86.4 & 86.4 & 0.693 & 0.803 & 55.2 & 85.7 & 85.7 & 0.529 & 0.766 & - & -\\
\rowcolor{gray!10} 
\textbf{Ours} & \textbf{53.6} & \textbf{88.2} & \textbf{87.6} & \textbf{0.579} & \textbf{0.817} & \textbf{58.4} & \textbf{87.1} & \textbf{87.1} & \textbf{0.477} & \textbf{0.791} & \textbf{75.3} & \textbf{72.2} \\
\bottomrule
\end{tabular}
}
\caption{Experimental results on CMU-MOSI, CMU-MOSEI, and MIntRec datasets. The results marked with $^\dagger$ are retrieved from \citet{Fang2025EMOE}, and those with $^\ddagger$ are cited from \citet{Chen2025MMA}.}
\label{tab:main_results}
\end{table*}

\subsection{Experimental Setup}
\subsubsection{Datasets and Feature Extraction}
To evaluate the robustness of our framework, we conduct experiments on four benchmarks: CMU-MOSI~\cite{Zadeh2016cmumosi}, CMU-MOSEI~\cite{bagherzadehetal2018cmumosei}, IEMOCAP~\cite{Busso2008IEMOCAPIE}, and MIntRec~\cite{MIntRec2022Zhang}.
Regarding feature extraction, we strictly adhere to the standard protocols from prior literature~\cite{Tsai2019MulT, Hazarika2020MISA}.
Detailed description, statistics and other information are provided in Appendix \ref{sec:appendix_datasets}.

% \subsubsection{Unified Imperfection Simulation}
% \label{sec:experimental_protocol}

% To rigorously evaluate the Reliability Spectrum Hypothesis (Sec.~\ref{sec:problem_formulation}), we adopt a unified simulation framework. \textbf{We operationalize the latent reliability $r_m$ through a controllable degradation coefficient $\lambda \in [0,1]$, establishing the inverse mapping $r_m \approx 1 - \lambda$.} Under this setting, $\lambda=0$ corresponds to the clean state ($r_m=1$), while $\lambda \to 1$ simulates the approach towards total information collapse ($r_m \to 0$).

% \paragraph{Evaluation Protocols.}
% Based on this simulation, we design three protocols to stress-test the model:
% \begin{itemize}
% \item \textbf{Protocol I: Modality Missingness.} Evaluates the boundary condition where $\lambda \to 1$. This tests "hard switching" capabilities when modalities are effectively unavailable.
% \item \textbf{Protocol II: Data Noise.} Evaluates the continuous interval $\lambda \in [0, 1]$ with global uniformity. This tests "soft filtering" capabilities against quality fluctuations.
% \item \textbf{Protocol III: Stochastic Mixture.} To benchmarks holistic robustness, we introduce a Static Heterogeneous Test Set. We independently sample $\lambda_m \sim \mathcal{U}[0, 1]$ for each modality of every test instance $x_i$, using a fixed random seed. This ensures that while the dataset reflects chaotic real-world distributions, the instance-level degradation remains identical for fair comparison.
% \end{itemize}

\subsubsection{Implementation Details}
We give brief information about the baselines, metrics and the implementation. And the detailed description is provided in Appendix~\ref{sec:appendix_experiment_setup}.

\noindent \textbf{Baselines.}
To verify the performance of our framework, we compare it against a comprehensive set of baselines categorized into two groups:

\noindent \textbf{(1) Standard Multimodal Learning}, under complete modalities, including TFN~\cite{Zadeh2017TensorFN}, LMF~\cite{liu2018efficient}, MulT~\cite{Tsai2019MulT}, MISA~\cite{Hazarika2020MISA}, MMIM~\cite{Han2021MMIM}, EMOE~\cite{Fang2025EMOE} and MMA~\cite{Chen2025MMA}.

\noindent \textbf{(2) Imperfect Multimodal Learning}, which is specifically designed for missing or noisy scenarios, including MulT~\cite{Tsai2019MulT}, MCTN~\cite{Pham2019MCTN}, MISA~\cite{Hazarika2020MISA}, MMIN~\cite{Zhao2021MMIN}, C-MIB \cite{Mai2023CMIB}, IMDER~\cite{Wang2024IMDER}, Multimodal-Boosting \cite{Mai2024MBoosting}, MoMKE~\cite{Xu2024MoMKE}, SAM-LML~\cite{mai2025supervised} and PaSE~\cite{Kang2025PaSE}.

\noindent \textbf{Evaluation Metrics.}
For CMU-MOSI and CMU-MOSEI, we follow ~\cite{Tsai2019MulT} to evaluate our method by using the metrics: 7-class Accuracy ($\text{ACC}_{7}$), Binary Accuracy ($\text{ACC}_{2}$), F1-score ($\text{F}_1$), and Mean-absolute Error (MAE). For MIntRec, we follow the standard protocol~\cite{Sun2024Contextual} to evaluate the results via: ACC and $\text{F}_1$. For IEMOCAP~\cite{liang2021multibench}, we use the average ACC and $\text{F}_1$ as evaluation metrics.

\noindent \textbf{Implementation Details.}
All models are implemented using PyTorch and trained on six NVIDIA RTX 4090 GPUs. We employ the Adam optimizer with a dropout rate of 0.1 to prevent overfitting. For QA-MoE, we construct the expert bank with $N=8$ GLU-based experts. The dual-path router is configured to activate the top-$k$ ($k=3$) experts for sparse computation.

\subsection{Performance on Standard Benchmarks}
\label{sec:benchmarks}
To ensure a fair evaluation of the architectural effectiveness, both the baselines and QA-MoE are trained on the original clean datasets without any degradation injection. Table \ref{tab:main_results} presents the performance comparison on aligned CMU-MOSI, CMU-MOSEI(unaligned is shown in Appendix~\ref{sec:appendix_perfect}) and MIntRec. On CMU-MOSI, our method surpasses the MMA by a improvement of \textbf{6.7\%} in $\text{ACC}_{7}$ and \textbf{1.2\%} in $\text{F}_{1}$ score. On other datasets, QA-MoE also outperforms prior methods consistently. It also indicates that the advantages are from the intrinsic design of our framework rather than data augmentation strategies.

\begin{table*}[t]
\centering
\renewcommand{\arraystretch}{1.2} %稍微增加行高，看着不挤
\resizebox{\textwidth}{!}{
\begin{tabular}{c|l|ccccccc|c}
\toprule
\multirow{2}{*}{\textbf{Datasets}} & \multirow{2}{*}{\textbf{Models}} & \multicolumn{7}{c|}{\textbf{Testing Condition (Available Modalities)}} & \multirow{2}{*}{\textbf{Avg.}} \\
\cmidrule(lr){3-9}
 & & $\{t\}$ & $\{a\}$ & $\{v\}$ & $\{t, a\}$ & $\{t, v\}$ & $\{a, v\}$ & $\{t, a, v\}$ & \\
\midrule

% ================= IEMOCAP Block =================
\multirow{5}{*}{\textbf{IEMOCAP}}
 & MulT & 62.4~/~63.7 & 49.7~/~51.6 & 48.9~/~45.7 & 68.3~/~69.4 & 67.8~/~68.3 & 56.3~/~55.8 & 70.1~/~70.5 & 60.5~/~60.7 \\
 & MISA & 66.5~/~68.0 & 56.5~/~59.0 & 52.5~/~51.6 & 72.9~/~75.1 & 72.6~/~73.6 & 63.9~/~65.4 & 74.2~/~74.5 & 65.5~/~66.7 \\
 & MMIM & 67.0~/~68.2 & 55.0~/~53.2 & 51.9~/~50.4 & 74.0~/~75.4 & 72.6~/~73.6 & 65.3~/~66.5 & 75.5~/~75.8 & 65.9~/~66.1 \\
 \rowcolor{gray!10} \cellcolor{white}
 & \textbf{Ours} & \textbf{71.2~/~72.3} & \textbf{58.2~/~59.1} & \textbf{54.6~/~53.8} & \textbf{75.1~/~75.3} & \textbf{73.8~/~74.1} & \textbf{66.9~/~66.8} & \textbf{77.1~/~77.3} & \textbf{68.1~/~68.4} \\
\midrule

% ================= CMU-MOSI Block =================
\multirow{7}{*}{\textbf{CMU-MOSI}}
 & MCTN  & 79.10~/~79.20 & 56.10~/~54.50 & 55.00~/~54.40 & 81.00~/~81.00 & 81.10~/~81.20 & 57.50~/~57.40 & 81.40~/~81.50 & 68.30~/~67.95 \\
 & MMIN  & 83.80~/~83.80 & 55.30~/~51.50 & 57.00~/~54.00 & 84.00~/~84.00 & 83.80~/~83.90 & 60.40~/~58.50 & 84.60~/~84.40 & 72.72~/~69.28 \\
 & IMDer  & 84.80~/~84.70 & 62.00~/~62.20 & 61.30~/~60.80 & 85.40~/~85.30 & 85.50~/~85.40 & 63.60~/~63.40 & 85.70~/~85.60 & 73.77~/~73.63 \\
 & MoMKE  & 86.59~/~86.52 & 63.19~/~58.61 & \textbf{63.35~/~63.34} & 87.20~/~87.17 & 87.04~/~87.00 & 64.04~/~64.66 & 87.96~/~87.89 & 75.24~/~74.55 \\
 & PaSE  & 84.70~/~84.23 & 60.01~/~58.79 & 61.43~/~61.50 & 86.71~/~86.79 & 87.14~/~86.99 & 63.35~/~63.32 & 88.32~/~88.25 & 73.89~/~73.60 \\
 \rowcolor{gray!10} \cellcolor{white}
 & \textbf{Ours} & \textbf{87.24~/~87.35} & \textbf{63.73~/~60.71} & 62.58~/~62.42 & \textbf{88.51~/~87.64} & \textbf{88.11~/~88.15} & \textbf{65.69~/~65.20} & \textbf{89.97~/~89.02} & \textbf{77.98~/~77.21} \\
\midrule

% ================= CMU-MOSEI Block =================
\multirow{7}{*}{\textbf{CMU-MOSEI}}
 & MCTN & 82.60~/~82.80 & 62.70~/~54.50 & 62.60~/~57.10 & 83.50~/~83.30 & 83.20~/~83.20 & 63.70~/~62.70 & 84.20~/~84.20 & 73.05~/~70.60 \\
 & MMIN & 82.30~/~82.40 & 58.90~/~59.50 & 59.30~/~60.00 & 83.70~/~83.30 & 83.80~/~83.40 & 63.50~/~61.90 & 84.30~/~84.20 & 71.92~/~71.75 \\
 & IMDer & 84.50~/~84.50 & 63.80~/~60.60 & 63.90~/~63.60 & 85.10~/~85.10 & 85.00~/~85.00 & 64.90~/~63.50 & 85.10~/~85.10 & 76.00~/~75.30 \\
 & MoMKE & 86.46~/~86.43 & 72.56~/~71.03& 70.12~/~70.23 & 86.68~/~86.61 & 86.79~/~86.69 & \textbf{73.34~/~71.82} & 87.12~/~87.03 & 79.33~/~78.80 \\
 & PaSE  & 84.36~/~84.08 & 69.04~/~68.56 & 68.69~/~68.74 & 86.47~/~86.42 & 86.73~/~86.45 & 72.03~/~71.90 & 88.10~/~87.96 & 77.89~/~77.69 \\
 \rowcolor{gray!10} \cellcolor{white}
 & \textbf{Ours} & \textbf{87.61~/~87.57} & \textbf{73.01~/~72.77} & \textbf{71.38~/~71.07} & \textbf{87.78~/~87.78} & \textbf{87.91~/~87.89} & 73.24~/~73.19 & \textbf{88.93~/~89.01} & \textbf{81.41~/~81.33} \\
\bottomrule
\end{tabular}
}
\caption{Performance comparison under various modality missingness scenarios. The values denote $\text{ACC}_{2} / \text{F}_{1}$. The testing conditions indicate the \textbf{available} modalities (e.g., $\{t\}$ means only Textual is available).}
\label{tab:fixed_missing_table}
\end{table*}

\subsection{Performance under Specific Imperfections}
\subsubsection{Evaluation on Protocol I}
\label{sec:missing_robustness}

To evaluate the model robustness against modality missingness, we adopt \textbf{Protocol I}. ACCording to ~\cite{Wang2024IMDER}, it has been divided into two commonly-used protocols.: 

\noindent \textbf{Fixed Missing Protocol.} It is designed to stimulate permanent sensor failure during inference. We test the models on all possible subsets of modalities obtained from the original modality dataset. The results in Table \ref{tab:fixed_missing_table} reveal that baseline models suffer great degradation without \textit{text}. In contrast, QA-MoE exhibits remarkable stability. Taking the $\{a, v\}$ setting as an example, the quality-aware router implicitly detects the missing text modality as having extreme aleatoric uncertainty. Consequently, the quality score $r_t$ asymptotically approaches zero, which automatically reduces the uninformative text branch. It proves that our model does not rely on a single modality but treats all modalities independently.

\noindent \textbf{Random Missing Protocol}. Consistent with~\cite{Lian2023GCNet}, we keep the same missing rate $\eta$ during training, validation, and testing phases. Table~\ref{tab:random_missing_table} reports the performance under varying missing rates $\eta$ from 10\% to 70\%. Compared with SAM-LML dropping 19.2\% on CMU-MOSI as $\eta$ shifts from 10\% to 70\%, QA-MoE achieves an average $\text{ACC}_{7}$ of \textbf{42.0\%}, which surpasses the strongest baseline by a substantial margin of \textbf{6.1\%}. It indicates that for each specific sample, our model can focus on the existing modalities, ensuring the high-fidelity inference in severe data sparsity.

\begin{table*}[t]
\centering
\renewcommand{\arraystretch}{1.1} %稍微增加行高，看着不挤
\resizebox{\textwidth}{!}{
\begin{tabular}{c|l|ccccccc|c}
\toprule
\multirow{2}{*}{\textbf{Dataset}} & \multirow{2}{*}{\textbf{Models}} & \multicolumn{7}{c|}{\textbf{Random Missing Rate ($\eta$)}} & \multirow{2}{*}{\textbf{Avg.}} \\
\cmidrule(lr){3-9}
 & & 10\% & 20\% & 30\% & 40\% & 50\% & 60\% & 70\% & \\
\midrule

% ================= CMU-MOSI Block (ACC-7 / F1) =================
\multirow{6}{*}{\rotatebox{90}{\textbf{CMU-MOSI}}} 
 & MCTN    & 39.8~/~78.5 & 38.5~/~75.7 & 35.5~/~71.2 & 32.9~/~67.6 & 31.2~/~64.8 & 29.7~/~62.5 & 27.5~/~59.0 & 33.6~/~68.5 \\
 & MMIN    & 41.2~/~81.8 & 38.9~/~79.1 & 36.9~/~76.2 & 34.9~/~71.6 & 34.2~/~66.5 & 29.1~/~64.0 & 28.4~/~61.0 & 34.5~/~71.5 \\
 & IMDer   & 42.1~/~83.4 & 41.6~/~80.5 & 37.4~/~77.6 & 35.2~/~66.3 & 29.5~/~65.4 & 27.0~/~65.5 & 26.5~/~60.4 & 34.2~/~71.3 \\
 & MoMKE   & 35.1~/~81.6 & 32.9~/~76.6 & 30.6~/~71.7 & 28.4~/~67.5 & 26.2~/~63.2 & 23.9~/~58.9 & 22.4~/~55.9 & 28.5~/~67.9 \\
 & SAM-LML & \underline{45.6~/~84.7} & \underline{42.9~/~81.2} & \underline{37.5~/~78.1} & \underline{37.8~/~74.7} & \underline{32.8~/~70.9} & \underline{28.1~/~66.6} & \underline{26.4~/~65.6} & \underline{35.9~/~74.5} \\
 % Ours Row
 \rowcolor{gray!10} \cellcolor{white}
 & \textbf{Ours}  & \textbf{53.3~/~85.1} & \textbf{51.2~/~81.5} & \textbf{47.2~/~78.4} & \textbf{40.5~/~76.4} & \textbf{37.2~/~73.2} & \textbf{33.9~/~69.1} & \textbf{30.5~/~68.7} & \textbf{42.0~/~76.1} \\
\midrule

% ================= CMU-MOSEI Block (ACC-7 / F1) =================
\multirow{6}{*}{\rotatebox{90}{\textbf{CMU-MOSEI}}} 
 & MCTN    & 49.8~/~81.6 & 48.6~/~78.7 & 47.4~/~76.2 & 45.6~/~74.1 & 45.1~/~72.6 & 43.8~/~71.1 & 43.6~/~70.5 & 46.3~/~75.0 \\
 & MMIN    & 50.6~/~81.3 & 49.6~/~78.8 & 48.1~/~75.5 & 47.5~/~72.6 & 46.7~/~70.7 & 45.6~/~70.3 & 44.8~/~69.5 & 47.6~/~74.1 \\
 & IMDer   & 52.1~/~82.9 & 51.3~/~79.7 & 49.6~/~77.8 & 48.0~/~73.3 & 46.6~/~68.4 & 45.0~/~65.9 & 44.1~/~66.6 & 48.1~/~73.5 \\
 & MoMKE   & 47.2~/~84.7 & 45.4~/~82.7 & 43.6~/~80.7 & 41.7~/~78.7 & 39.8~/~76.7 & 37.9~/~74.7 & 36.7~/~73.3 & 41.8~/~78.8 \\
 & SAM-LML & \underline{51.9~/~84.5} & \underline{51.7~/~83.7} & \underline{48.7~/~81.6} & \underline{48.3~/~79.5} & \underline{46.9~/~77.4} & \underline{45.6~/~76.4} & \underline{44.5~/~74.6} & \underline{48.2~/~79.7} \\
 % Ours Row
 \rowcolor{gray!10} \cellcolor{white}
 & \textbf{Ours}  & \textbf{55.8~/~86.4} & \textbf{54.2~/~81.9} & \textbf{51.2~/~80.3} & \textbf{50.3~/~79.9} & \textbf{48.3~/~78.7} & \textbf{47.4~/~77.0} & \textbf{46.1~/~75.3} & \textbf{50.5~/~79.9} \\
\bottomrule
\end{tabular}
}
\caption{Robustness comparison under \textbf{Random Missing Protocol} ($\text{ACC}_7 / \text{F}_1)$.}
\label{tab:random_missing_table}
\end{table*}

\begin{table}[t]
\centering
\renewcommand{\arraystretch}{1.1} % 增加行高，让表格更舒展
\resizebox{\columnwidth}{!}{ % 自动调整宽度适应文档
\begin{tabular}{c|c|ccc|c}
\toprule
\multirow{2}{*}{\textbf{Dataset}} & \multirow{2}{*}{\textbf{NR}} & \textbf{C-MIB} & \textbf{MM-Boosting} & \textbf{SAM-LML} & \cellcolor{gray!10}\textbf{QA-MoE (Ours)} \\
 & &  $\text{ACC}_{2}$ / MAE &  $\text{ACC}_{2}$/ MAE &  $\text{ACC}_{2}$ / MAE & \cellcolor{gray!10}\textbf{ $\text{ACC}_{2}$ / MAE} \\
\midrule

% ===================== MOSI BLOCK =====================
\multirow{8}{*}{\rotatebox{90}{\textbf{CMU-MOSI}}} 
 & 0.1 & 87.8 / 0.670 & 86.7 / 0.678 & 88.4 / 0.636 & \cellcolor{gray!10}\textbf{89.4 / 0.616} \\
 & 0.2 & 87.5 / 0.726 & 86.1 / 0.738 & 88.1 / 0.665 & \cellcolor{gray!10}\textbf{88.9 / 0.636} \\
 & 0.3 & 86.4 / 0.912 & 86.4 / 0.785 & 87.8 / 0.663 & \cellcolor{gray!10}\textbf{88.6 / 0.639} \\
 & 0.4 & 83.2 / 1.366 & 85.5 / 0.841 & 87.6 / 0.666 & \cellcolor{gray!10}\textbf{88.4 / 0.641} \\
 & 0.5 & 84.9 / 1.660 & 86.1 / 1.172 & 88.1 / 0.666 & \cellcolor{gray!10}\textbf{88.1 / 0.649} \\
 & 0.6 & 80.8 / 2.595 & 82.0 / 1.355 & 87.5 / 0.660 & \cellcolor{gray!10}\textbf{87.8 / 0.652} \\
 & 0.7 & 82.1 / 3.146 & 84.4 / 1.750 & 87.3 / 0.669 & \cellcolor{gray!10}\textbf{87.7 / 0.660} \\
\cmidrule(lr){2-6}
 & \textbf{Avg.} & 84.7 / 1.582 & 85.3 / 1.046 & 87.8 / 0.661 & \cellcolor{gray!10}\textbf{88.4 / 0.642} \\
\midrule

% ===================== MOSEI BLOCK =====================
\multirow{8}{*}{\rotatebox{90}{\textbf{CMU-MOSEI}}} 
 & 0.1 & 86.1 / 0.545 & 86.4 / 0.544 & 87.0 / 0.521 & \cellcolor{gray!10}\textbf{88.2 / 0.498} \\
 & 0.2 & 84.5 / 0.582 & 86.6 / 0.557 & 87.0 / 0.525 & \cellcolor{gray!10}\textbf{88.1 / 0.501} \\
 & 0.3 & 85.6 / 0.622 & 85.5 / 0.623 & 87.3 / 0.522 & \cellcolor{gray!10}\textbf{88.0 / 0.512} \\
 & 0.4 & 84.4 / 0.703 & 85.3 / 0.682 & 87.2 / 0.525 & \cellcolor{gray!10}\textbf{87.6 / 0.514} \\
 & 0.5 & 83.7 / 0.875 & 84.1 / 0.724 & 86.6 / 0.529 & \cellcolor{gray!10}\textbf{87.4 / 0.521} \\
 & 0.6 & 82.4 / 1.054 & 85.4 / 0.924 & 87.0 / 0.532 & \cellcolor{gray!10}\textbf{87.2 / 0.523} \\
 & 0.7 & 80.5 / 1.404 & 80.3 / 1.125 & 85.9 / 0.545 & \cellcolor{gray!10}\textbf{86.8 / 0.531} \\
\cmidrule(lr){2-6}
 & \textbf{Avg.} & 83.9 / 0.826 & 84.8 / 0.740 & 86.9 / 0.528 & \cellcolor{gray!10}\textbf{87.6 / 0.514} \\
\bottomrule
\end{tabular}
}
\caption{Comparison under varying NI~$(\lambda)$.The results are cited from \cite{mai2025supervised}.}
\label{tab:noise_robustness_comparison}
\end{table}

\begin{table}[htbp]
\centering
\resizebox{1.0\columnwidth}{!}{
\begin{tabular}{l|c|cccc}
\toprule
\multirow{2}{*}{\textbf{Model}} & \multirow{2}{*}{\textbf{Training Strategy}} & \multicolumn{4}{c}{\textbf{Mixed Test Set (Protocol III)}} \\
 & & MAE $\downarrow$ & $\text{ACC}_{7}$ $\uparrow$ & $\text{ACC}_{2} \uparrow$ & $\text{F}_{1} \uparrow$ \\
\midrule
\multirow{2}{*}{MMA} & Clean Train & 0.693 & 46.9 & 86.4 & 86.4 \\
 & \cellcolor{gray!10}Spectrum Train & \cellcolor{gray!10}0.688 & \cellcolor{gray!10}51.8 & \cellcolor{gray!10}87.2 & \cellcolor{gray!10}88.4 \\
\midrule
\multirow{2}{*}{SAM-LML} & Clean Train & 0.628 & 49.4 & 89.2 & 89.1 \\
 & \cellcolor{gray!10}Spectrum Train & \cellcolor{gray!10}0.599 & \cellcolor{gray!10}52.5 & \cellcolor{gray!10}89.6 & \cellcolor{gray!10}89.7 \\
\midrule
\multirow{2}{*}{QA-MoE (Ours)} & Clean Train & 0.589 & 53.6 & 87.3 & 87.6 \\
 & \cellcolor{gray!10}Spectrum Train & \cellcolor{gray!10}\textbf{0.515} & \cellcolor{gray!10}\textbf{54.5} & \cellcolor{gray!10}\textbf{87.9} & \cellcolor{gray!10}\textbf{89.1} \\
\bottomrule
\end{tabular}
}
\caption{Decoupling analysis under Protocol III.}
\label{tab:decoupling}
\end{table}

\subsubsection{Evaluation on Protocol II}
\label{sec:noise_robustness}

To comprehensively evaluate model robustness against modality noise, we adopt the diverse noise injection protocol established in \cite{mai2025supervised}. The Noise Intensity~$\lambda$ varies from $0.1$ to $0.7$, controlling the intensity of the degradation. We compare QA-MoE against state-of-the-art robust frameworks using their reported settings. Table \ref{tab:noise_robustness_comparison} visualizes the performance trends on CMU-MOSI and CMU-MOSEI. As the noise intensity increases, standard baselines exhibit rapid performance decay.  While SAM-LML shows improved resistance, QA-MoE consistently outperforms all baselines across the entire noise spectrum.  Notably, at severe noise levels ($\lambda=0.7$), QA-MoE maintains a lead of roughly \textbf{0.9\%} over SAM-LML, which validates that our distributional reliability scoring effectively filters out high-variance features and reconstructs semantics from noisy signals.

\subsection{Performance on Spectrum Dataset}
% \subsection{Evaluation on Protocal III}
\label{sec:unified_evaluation}

Moving beyond standard benchmarks, we evaluate the \textbf{Protocol III} under the proposed Continuous Reliability Spectrum. The model encounters a heterogeneous test set comprising random combinations of noise ($\lambda$) and missingness ($\eta$), simulating the unpredictable imperfections of real-world deployment. 

\subsubsection{Analysis of Architectural Superiority}
\label{sec:architectural_superiority}
A fundamental question arises regarding the source of our model's efficacy: \textit{Does the robustness of QA-MoE stem from its intrinsic architectural design, or merely from the Spectrum-Aware training strategy?} To rigorously disentangle these factors, we conduct a controlled experiment by retraining the strongest baselines using the exact same dynamic degradation injection strategy employed in our framework. Table~\ref{tab:decoupling} reports the performance on the heterogeneous Protocol III test set. Remarkably, the QA-MoE trained solely on clean data (53.6\%) still outperforms the strongest baseline enhanced by the spectrum training strategy (52.5\%). It convinces that our robustness derives primarily from the proposed mechanism in handling unseen shifts, rather than relying on data augmentation.

\subsubsection{"One-Checkpoint-for-All" Strategy}
\label{sec:one_checkpoint}

\begin{figure}[t]
    \centering
    \includegraphics[width=0.95\columnwidth]{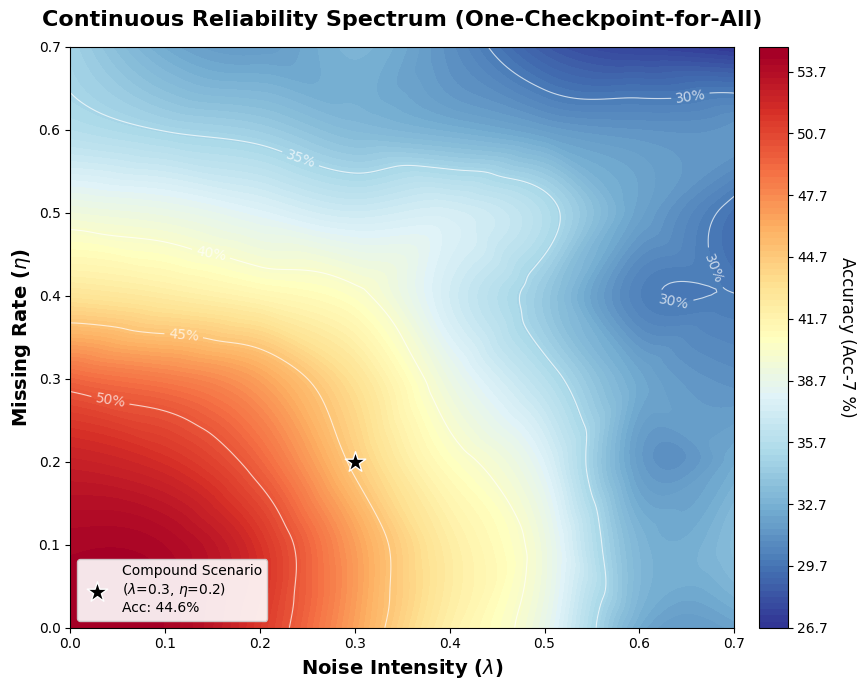} 
    \caption{\textbf{Continuous Reliability Landscape.} The smooth performance gradient (from warm to cool colors) demonstrates that QA-MoE exhibits graceful degradation rather than abrupt failure. The star ($\star$) marks the compound defect scenario ($\lambda=0.3, \eta=0.2$).}
    \label{fig:landscape}
\end{figure}

Unlike existing methods often require retraining to adapt to specific noise levels, QA-MoE is designed to remain effective across varying reliability conditions. To verify this, we evaluate a single trained checkpoint across the entire reliability spectrum grid without any parameter tuning.
Figure~\ref{fig:landscape} visualizes the model's performance on the Continuous Reliability Spectrum introduced in Figure~\ref{fig:spectrum}. Unlike discrete evaluations, this continuous surface demonstrates a high-fidelity plateau covering the Stochastic Mixture region. The model maintains robustness well into the degradation zones ($\star$), confirming its ability to adapt continuously to unseen defects. The detailed value of each discrete points are also provided in Appendix~\ref{sec:appendix_spectrum}. Besides, it also provides the performance under other SAM-LML, which shows that our model can maintain effective across the spectrum without retraining.

\subsection{Model Analysis}
In this section, we conduct a comprehensive analysis to provide deeper insights into the properties of QA-MoE. Besides the following, the analysis of the computational efficiency is placed in Appendix~\ref{sec:appendix_efficiency} due to space.

% The analysis as follows is based on the CMU-MOSI dataset, and the others is shown in Appendix~\ref{sec:appendix_efficiency}. 

\begin{table}[htbp]
\centering
\resizebox{1.0\columnwidth}{!}{
\begin{tabular}{l|cccc}
\toprule
\multirow{2}{*}{\textbf{Model Variants}} & \multicolumn{4}{c}{\textbf{Spectrum Training Setting}} \\
 & MAE $\downarrow$ & $\text{ACC}_{7} \uparrow$ & $\text{ACC}_{2} \uparrow$ & $\text{F}_1 \uparrow$ \\ \midrule
\textbf{QA-MoE (Full)} & \textbf{0.525} & \textbf{54.5} & \textbf{88.8} & \textbf{89.1} \\ \midrule
\ w/o Quality Gating & 0.884 & 52.1 & 76.2 & 77.4 \\
\ w/o Variance ($\boldsymbol{\sigma}^2$) & 0.795 & 50.7 & 79.1 & 51.2 \\
\ w/o Universal Fallback ($\mathbf{y}_{prior}$) & 0.780 & 48.6 & 78.5 & 45.9 \\
\bottomrule
\end{tabular}
}
\caption{Ablation studies on CMU-MOSI. We evaluate the contribution of the key parts of our QA-MoE.}
\label{tab:ablation}
\end{table}

\noindent \textbf{Ablation Study.}
\label{sec:ablation}
To disentangle component contributions, we conduct ablation studies on CMU-MOSI (Table \ref{tab:ablation}) with Spectrum Training. First, removing the quality gate significantly increases errors on noisy inputs, confirming the necessity of explicit signals to bypass unreliable experts. Second, relying solely on the mean vector causes performance degradation, validating that the second moment is a critical proxy for aleatoric uncertainty.

\begin{figure}[t]
    \centering
    \includegraphics[width=\linewidth]{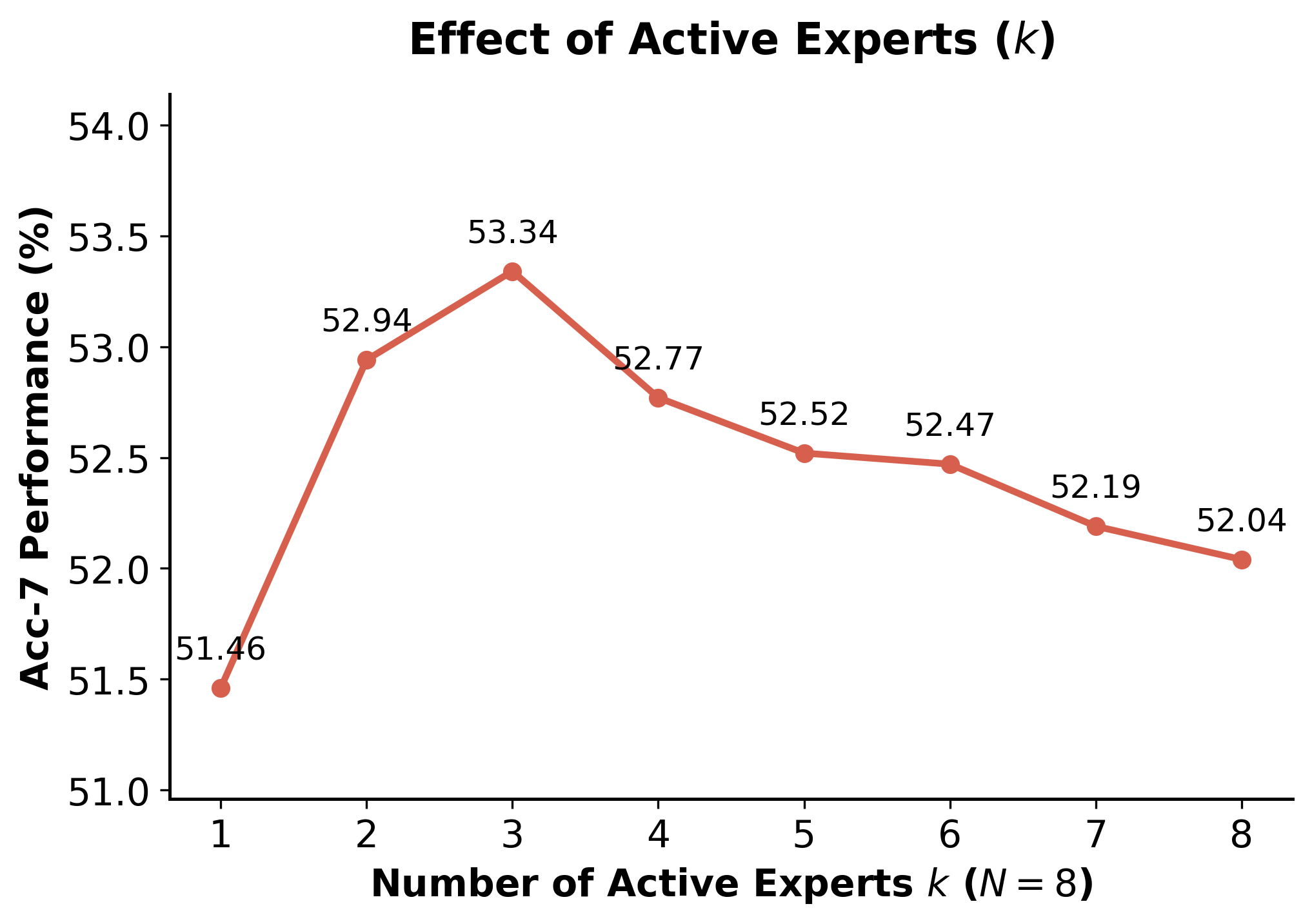}
    \caption{Parameter $k$ Sensitivity Analysis.} 
    \label{fig:sensitivity}
\end{figure}

\noindent \textbf{Parameter Sensitivity Analysis.}
\label{sec:sensitivity} We conduct sensitivity analysis on the number of active experts ($k$). We fix the total number of experts $N=8$ and vary the active selection $k \in \{1, 2, 3, 4, 8\}$. Figure~\ref{fig:sensitivity} shows a performance peak at $k = 3$. The tendency indicates that a single expert is insufficient to capture complex multimodal dynamics and increasing $k$ to 8 results in degradation due to overfitting. Thus, $k=3$ represents the optimal trade-off between effectiveness and efficiency.

\begin{figure}[t]
    \centering
    \includegraphics[width=1.0\linewidth]{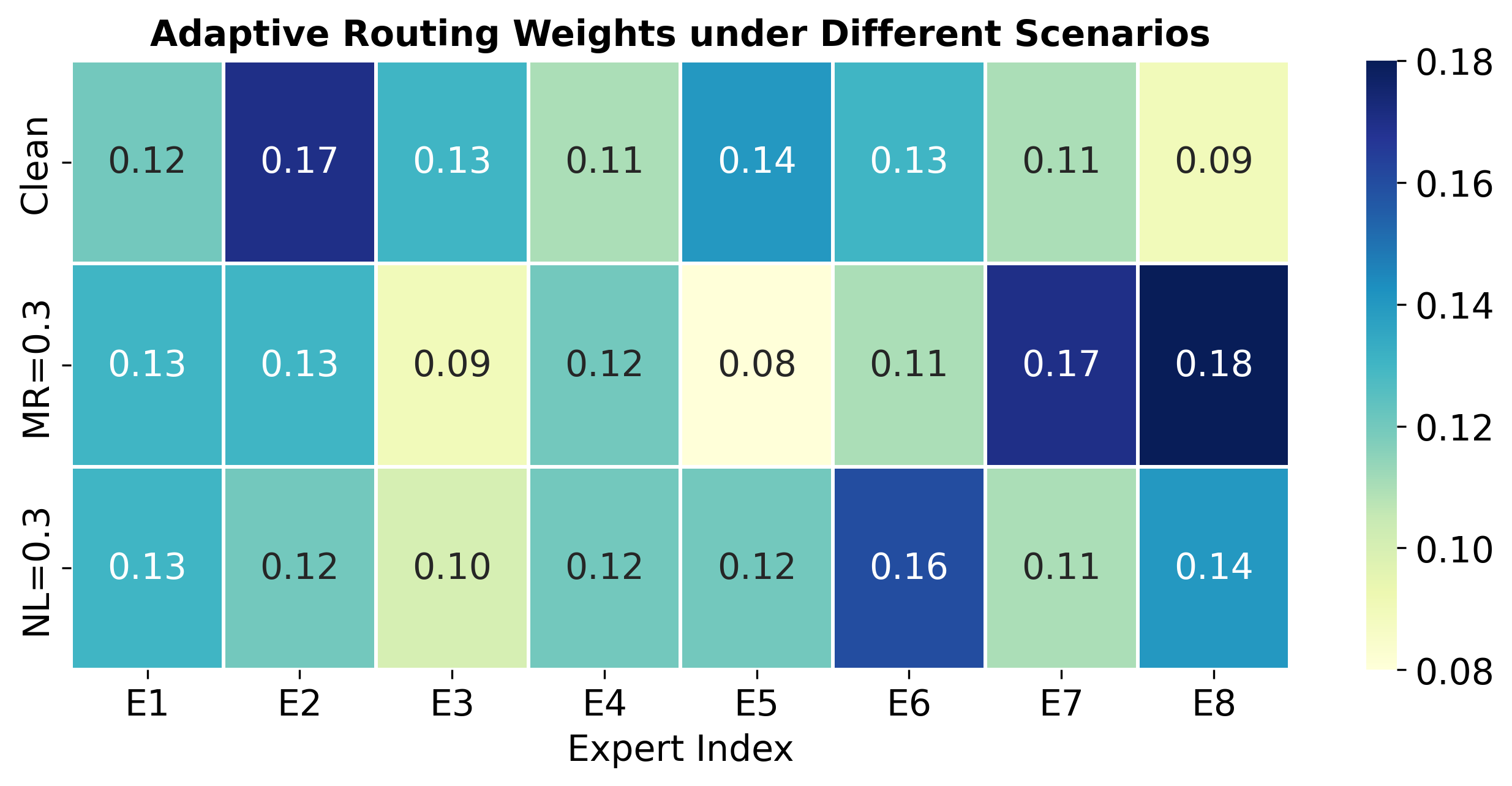}
    \caption{Visualization of Adaptive Routing Patterns.} 
    \label{fig:routing_vis}
\end{figure}

\noindent \textbf{Interpretability Analysis.}
\label{sec:interpretability_analysis} Figure~\ref{fig:routing_vis} shows the reconfiguration of expert attention under varying degradation contexts. We observe a distinct quality-aware routing shift. Notably, Expert 2 dominates in clean settings but is suppressed under corruption to prevent error propagation. In contrast, Expert 8 gains prominence specifically under imperfect scenarios. Besides, the divergent behavior in Expert 5 under different conditions confirms that QA-MoE discriminates between specific failure modes rather than applying a generic penalty.

\section{Conclusion}
In this work, we introduce the \textbf{Continuous Reliability Spectrum} to model real-world imperfections and propose \textbf{QA-MoE} to address them. By dynamically routing signals based on quality, QA-MoE achieves a \textbf{"One-Checkpoint-for-All"} capability across synthetic protocols. A risk is the effectiveness of the routing relies on the precision of quality predictors, where estimation errors in extreme edge cases could lead to suboptimal performance.

\section*{Limitations}
Despite the effectiveness of our approach, there are two main limitations. First, the quality signals in our framework are learned implicitly in a self-supervised manner; thus, the model lacks explicit interpretability regarding specific noise types (e.g., blur vs. occlusion). Second, the MoE architecture involves a routing mechanism that introduces additional computational overhead compared to static fusion methods. We plan to explore more efficient routing strategies and fine-grained quality modeling in future work.

% Bibliography entries for the entire Anthology, followed by custom entries
%\bibliography{anthology,custom}
% Custom bibliography entries only
\bibliography{custom}

\clearpage
\appendix

\section{Appendix}
\label{sec:appendix}

\subsection{Spectrum Dataset Generation Details}
\label{sec:appendix_spectrum_generation}

To implement the \textbf{Spectrum-Aware Training Strategy} described in Section~3.5, we construct a dynamic data loader that applies stochastic transformations on-the-fly. Unlike static augmentation, this process generates a unique view of the dataset for every training batch, ensuring the model traverses the continuous reliability spectrum.

\subsubsection{Dynamic Injection Protocol}
To ensure the model generalizes across the entire reliability spectrum, we employ a batch-wise dynamic sampling strategy.
Specifically, at the beginning of each training iteration, we sample a pair of global degradation coefficients $(\lambda_{batch}, \eta_{batch})$ from a uniform distribution:
\begin{equation}
\lambda_{batch} \sim \mathcal{U}(0, 1), \quad \eta_{batch} \sim \mathcal{U}(0, 1)
\end{equation}
These coefficients are then applied uniformly to all samples within the current mini-batch. This exposes the router to continuously varying difficulty levels throughout the training epochs, preventing overfitting to any specific discrete noise intensity.

\subsubsection{Modality-Specific Degradation}
Since the modalities (Text, Audio, Vision) have distinct physical properties, we design specific degradation functions $\mathcal{T}(\cdot)$ for each, consistent with the Stochastic Imperfection Modeling in Eq.~\ref{eq:transformation}.

\paragraph{Continuous Modalities (Audio \& Vision).}
For the continuous feature vectors from acoustic (e.g., COVAREP/Wav2Vec) and visual (e.g., Facet/ViT) encoders, we verify robustness by injecting additive noise. The corrupted feature $\tilde{\mathbf{u}}_m$ is generated as:
\begin{equation}
    \tilde{\mathbf{u}}_m = \mathbf{u}_m + \boldsymbol{\epsilon}_m, \quad \boldsymbol{\epsilon}_m \sim \mathcal{N}(\mathbf{0}, (\lambda \cdot \sigma_{\text{ref}})^2 \mathbf{I})
\end{equation}
where $\sigma_{\text{ref}}$ is a reference standard deviation calculated from the training set statistics to ensure the noise scale is relative to the feature.

\noindent \textbf{Acoustic:} We simulate background noise and sensor jitter using Additive White Gaussian Noise (AWGN).

\noindent \textbf{Visual:} We simulate blur and low-light sensor noise. While actual blur is a convolution operation, in the high-level feature space, this is effectively modeled by increasing the feature variance via additive Gaussian noise.

\paragraph{Discrete Modality (Text).}
For textual data, noise manifests as Automatic Speech Recognition (ASR) errors or missing words. We implement this via a \textbf{Token-Level Dropout} mechanism. Given a sequence of word embeddings $\mathbf{u}_t = \{w_1, w_2, \dots, w_L\}$, each token is independently replaced by a zero vector (or a special \texttt{[MASK]} token) with probability $p = \lambda$:
\begin{equation}
    \tilde{w}_i = 
    \begin{cases} 
    w_i & \text{with probability } 1 - \lambda \\
    \mathbf{0} & \text{with probability } \lambda 
    \end{cases}
\end{equation}
This simulates semantic fragmentation ranging from minor typos (low $\lambda$) to unreadable sentences (high $\lambda$).

\paragraph{Modality Missingness.}
Finally, to simulate complete sensor failure (Protocol I), we apply the global missingness mask $\mathbb{I}_{\mathrm{miss}}$. With probability $\eta$, the entire feature sequence for a modality $m$ is zeroed out: $\tilde{\mathbf{u}}_m \leftarrow \mathbf{0}$.

\label{sec:appendix_standard}
\begin{table*}[t]
\centering
\renewcommand{\arraystretch}{0.6}
\resizebox{0.95\textwidth}{!}{
\begin{tabular}{l|ccccc|ccccc}
\toprule
\multirow{2}{*}{\textbf{Models}} & \multicolumn{5}{c|}{\textbf{CMU-MOSI}} & \multicolumn{5}{c}{\textbf{CMU-MOSEI}} \\
 & ACC-7 $\uparrow$ & ACC-2 $\uparrow$ & F1 $\uparrow$ & MAE $\downarrow$ & Corr $\uparrow$ & ACC-7 $\uparrow$ & ACC-2 $\uparrow$ & F1 $\uparrow$ & MAE $\downarrow$ & Corr $\uparrow$ \\
\midrule
TFN$^\dagger$   & 35.3 & 76.5 & 76.6 & 0.995 & 0.698 & 50.2 & 84.2 & 84.0 & 0.573 & 0.700 \\
LMF$^\dagger$   & 31.1 & 79.1 & 79.1 & 0.963 & 0.695 & 51.9 & 83.8 & 83.9 & 0.565 & 0.677 \\
MulT$^\dagger$  & 33.2 & 80.3 & 80.3 & 0.933 & 0.711 & 53.2 & 84.0 & 84.0 & 0.556 &   -   \\
MISA$^\dagger$  & 43.6 & 83.8 & 83.9 & 0.742 & 0.761 & 51.0 & 84.8 & 84.8 & 0.557 & 0.756 \\
MMIM$^\dagger$  & 45.9 & 83.4 & 83.4 & 0.777 &   -   & 52.6 & 81.5 & 81.3 & 0.578 &   -   \\
EMOE$^\dagger$  & 47.8 & 85.4 & 85.3 & 0.697 &   -   & 53.9 & 85.5 & 85.5 & 0.530 &   -   \\
\rowcolor{gray!10} 
\textbf{Ours} & \textbf{53.3} & \textbf{87.4} & \textbf{87.2} & \textbf{0.583} & \textbf{0.816} & \textbf{58.4} & \textbf{87.1} & \textbf{87.1} & \textbf{0.477} & \textbf{0.791} \\
\bottomrule
\end{tabular}
}
\caption{Experimental results on \textbf{CMU-MOSI} and \textbf{CMU-MOSEI} datasets. The results marked with $^\dagger$ are retrieved from \citet{Fang2025EMOE}, and those with $^\ddagger$ are cited from \citet{Chen2025MMA}.}
\label{tab:appendix_cmu_results}
\end{table*}

\begin{table*}[t]
\centering
\renewcommand{\arraystretch}{0.6} % 增加行高，让表格更好看
\resizebox{0.90\textwidth}{!}{
\begin{tabular}{l|cccccccc}
\toprule
\multirow{2}{*}{\textbf{MR} ($\eta$)} & \multicolumn{8}{c}{\textbf{Noise Intensity} ($\lambda$)} \\
\cmidrule(lr){2-9}
 & 0.0 & 0.1 & 0.2 & 0.3 & 0.4 & 0.5 & 0.6 & 0.7 \\
\midrule
0\%   & \textbf{54.50} & 54.27 & 51.31 & 46.65 & 41.98 & 38.63 & 32.36 & 31.92 \\
10\%          & 54.20 & 53.98 & 51.60 & 46.79 & 42.27 & 39.07 & 32.94 & 33.38 \\
20\%          & 52.60 & 51.79 & 49.27 & \cellcolor{gray!20}\textbf{44.61} & 40.67 & 37.76 & 31.63 & 32.36 \\
30\%          & 49.25 & 48.31 & 46.65 & 43.15 & 38.92 & 35.71 & 32.22 & 31.05 \\
40\%         & 43.06 & 42.38 & 41.44 & 40.38 & 37.17 & 34.11 & 30.32 & 30.03 \\
50\%         & 39.31 & 38.85 & 37.90 & 36.94 & 37.32 & 35.57 & 31.92 & 29.59 \\
60\%         & 35.69 & 34.94 & 34.40 & 33.23 & 32.59 & 31.71 & 31.20 & 31.34 \\
70\% & 34.69 & 32.34 & 31.55 & 32.80 & 31.03 & 28.53 & 27.47 & 26.88 \\
\bottomrule
\end{tabular}
}
\caption{\textbf{Discrete Evaluation Grid (ACC-7 \%).} This table presents the exact performance metrics of the single QA-MoE checkpoint across varying degrees of degradation. The gray cell marks the compound defect scenario ($\lambda=0.3, \eta=20\%$) analyzed in Figure~\ref{fig:landscape}.}
\label{tab:spectrum_grid}
\end{table*}

\subsection{Datasets and Feature Extraction}
\label{sec:appendix_datasets}
\textbf{CMU-MOSI}~\cite{Zadeh2016cmumosi} and \textbf{CMU-MOSEI}~\cite{bagherzadehetal2018cmumosei} are the most widely used benchmarks for MSA and MER tasks. CMU-MOSI consists of 2,199 opinion video clips labeled with sentiment intensity scores ranging from -3 (highly negative) to +3 (highly positive). CMU-MOSEI is a larger-scale dataset containing 23,453 annotated video segments. Both datasets are pre-processed and word-aligned following the standard protocol. We strictly follow the standard feature extraction protocols established in prior literature~\cite{Tsai2019MulT,Hazarika2020MISA}, and we utilize 300-dimensional GloVe language features~\cite{pennington2014glove} and 768-dimensional BERT-base-uncased hidden states~\cite{Devlin2019BERTPO}. Facet~\cite{BaltrušaitisOpenFace2016} provides 35 facial action unit visual features, and COVAREP~\cite{Degottex2014COVAREP} offers 74-dimensional acoustic features.

\noindent \textbf{IEMOCAP}~\cite{Busso2008IEMOCAPIE} is a multimodal database for emotion recognition, comprising dyadic conversations between ten speakers. Following prior works~\cite{Tsai2019MulT}, we focus on the classification of six discrete emotions: happy, sad, angry, fearful, frustrated, and neutral. For IEMOCAP, we follow~\cite{Zhao2021MMIN} to extract acoustic, visual and textual features.

\noindent \textbf{MIntRec}~\cite{MIntRec2022Zhang} is a challenging dataset for multimodal intent recognition capturing high-quality "in-the-wild" interactions. Unlike lab-controlled datasets, MIntRec naturally contains environmental noise and diverse background scenes, making it an ideal testbed for evaluating model robustness against real-world imperfections. On MIntRec, dimensions for text, visual, and acoustic features are 768, 256, and 768, respectively.

\subsection{Baslines, Implementation and Metrics}
\label{sec:appendix_experiment_setup}

\textbf{Baselines}. To verify the effectiveness of our framework, we compare it against a comprehensive set of baselines categorized into two groups:

General Multimodal Learning Approaches: We select methods that focus on sophisticated fusion mechanisms assuming complete modalities. These include TFN~\cite{Zadeh2017TensorFN} and LMF~\cite{liu2018efficient} which utilize tensor fusion; MulT~\cite{Tsai2019MulT} which employs cross-modal transformers; MISA~\cite{Hazarika2020MISA} which focuses on feature disentanglement; MMIM~\cite{Han2021MMIM} which maximizes mutual information; and inspired by Mixture of Experts, EMOE~\cite{Fang2025EMOE} and MMA~\cite{Chen2025MMA} facilitate adaptive multimodal fusion.

Robustness-Oriented Approaches: We also compare against methods specifically designed for handling missing or noisy modalities. These include MMIN~\cite{Zhao2021MMIN}which reconstructs missing modalities via cascaded prediction, and SMIL~\cite{Ma2021SMILML} which utilizes Bayesian meta-learning to handle severe modality absence.

\noindent \textbf{Implementation}.
We implement all models using PyTorch on NVIDIA RTX 4090 GPUs. Following standard protocols~\citep{Tsai2019MulT}, we utilize BERT-base-uncased ($d_t=768$) for text, and acoustic/visual features extracted via COVAREP ($d_a=74$) and Facet ($d_v=35$), respectively. The hidden dimension of the multimodal encoders is set to $d_{model}=128$. Models are trained for 30 epochs with a batch size of 16. We employ the Adam optimizer with $\beta=(0.9, 0.999)$ and a weight decay of $1e^{-5}$. To prevent overfitting, we apply a dropout rate of 0.1 and gradient clipping (threshold 1.0). The learning rate is tuned via grid search within $\{1e^{-3}, 5e^{-4}, 1e^{-4}\}$ and decayed using a Cosine Annealing scheduler. For the QA-MoE structure, we set the total experts $N=8$ and active experts $k=3$. The Quality Predictors are implemented as two-layer MLPs to ensure lightweight computation. For strict reproducibility, all experiments are conducted with a fixed random seed (1111).

\subsection{Supplementary Experimental Results}
\subsubsection{Results on Perfect Dataset}
\label{sec:appendix_perfect}
We evaluate the model under the challenging unaligned setting, where modalities possess inherent temporal asynchrony. Compared to strong baselines, including the recent MoE-based method EMOE~\citep{Fang2025EMOE}, QA-MoE achieves substantial gains across all metrics.
On \textbf{CMU-MOSI}, we achieve an $\text{ACC}_{7}$ of \textbf{53.3\%}, surpassing the previous best (EMOE) by \textbf{+5.5\%}. 
On \textbf{CMU-MOSEI}, we get an $\text{ACC}_{7}$ of \textbf{58.4\%}, outperforming the strongest baseline by \textbf{+4.5\%}. 
These results confirm that our Quality-Aware Routing mechanism is not solely a defensive measure against noise. Even in perfect datasets, the router effectively dynamically selects experts to handle the natural semantic misalignment and heterogeneity inherent in unaligned multimodal settings which proves the architecture's intrinsic superiority.

\subsubsection{Results on Spectrum Dataset}
\label{sec:appendix_spectrum}
To ensure reproducibility and transparency, we provide the comprehensive numerical results corresponding to the One-Checkpoint-for-All evaluation discussed in Section~4.3.2. Table~\ref{tab:spectrum_grid} details the $\text{ACC}_7$ performance of QA-MoE across the complete discrete grid of Noise Intensities ($\lambda \in [0, 0.7]$) and Missing Rates ($\eta \in [0, 70\%]$). These raw values serve as the basis for the continuous landscape visualization in Figure~\ref{fig:landscape}.
Notably, the table confirms the model's graceful degradation:

(1) Under \textbf{Ideal Conditions} ($\lambda=0, \eta=0$), the model achieves a peak accuracy of \textbf{54.50\%}.

(2) Under the \textbf{Compound Scenario} highlighted in the main text ($\lambda=0.3, \eta=20\%$), the model retains a robust accuracy of \textbf{44.61\%}, validating the effectiveness of the quality-aware routing mechanism even when subject to simultaneous mixed imperfections.

\begin{figure}[t] 
    \centering 
    \includegraphics[width=0.95\columnwidth]{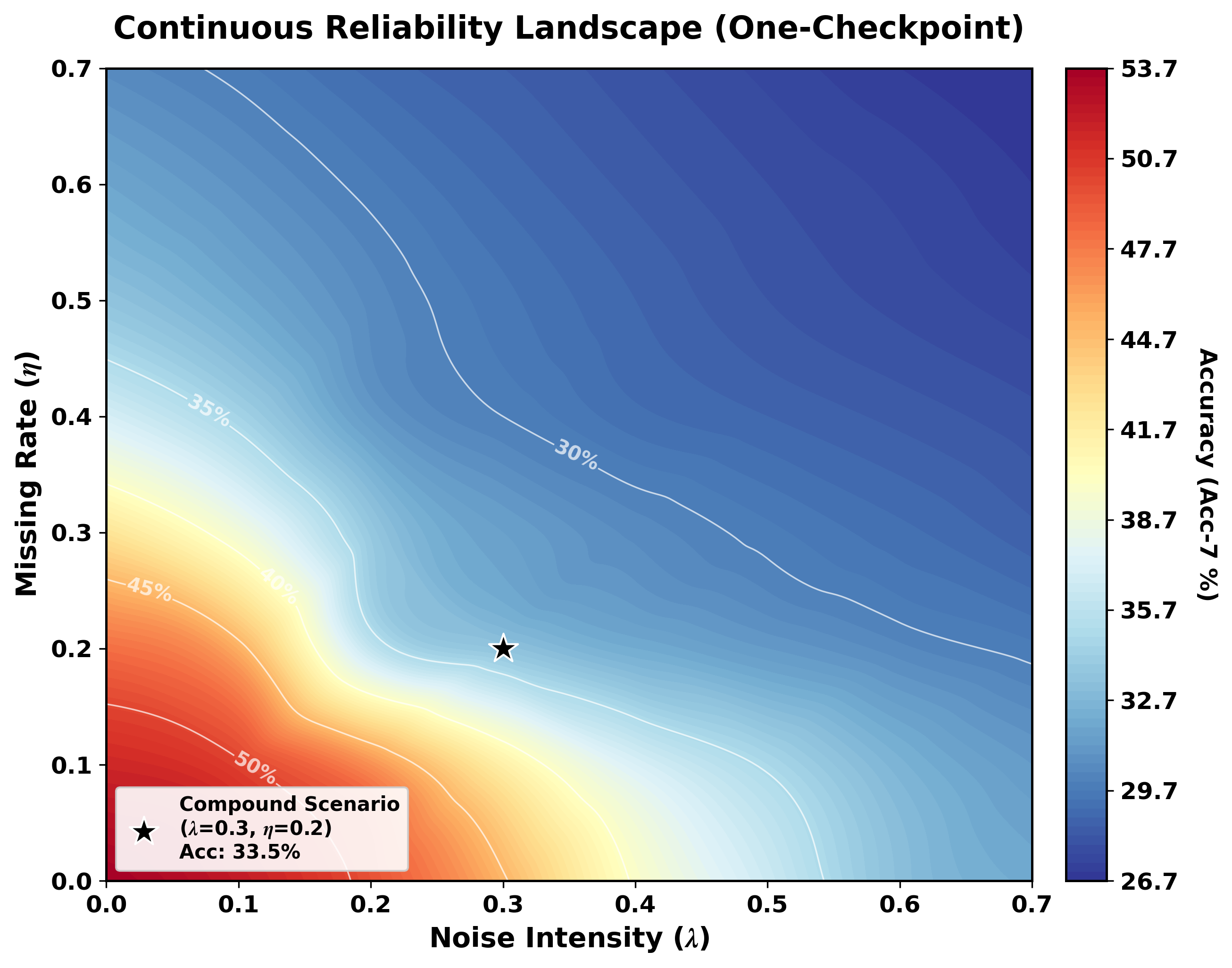} 
    \caption
    {\textbf{Reliability Landscape of Baseline (SAM-LML).} The visualization reveals a sharp performance decay, forming a reliability cliff.'' While the model achieves peak performance at the clean origin, its accuracy plummets rapidly as degradation intensity increases. The star ($\star$) marks the compound defect scenario ($\lambda=0.3, \eta=0.2$), where accuracy has already degraded to 35.5\%, demonstrating the lack of robustness in the One-Checkpoint'' setting.} 
    \label{fig:sam_lml_cliff} 
\end{figure}

To contrast with the stability of QA-MoE, we visualize the reliability landscape of the strongest baseline, SAM-LML, in Figure~\ref{fig:sam_lml_cliff}. Unlike our proposed method, which maintains a high-performance plateau, SAM-LML exhibits significant \textbf{brittleness} to unseen degradation.As observed, the high-accuracy region (red/orange) is strictly confined to the top-left corner (clean data). A minor shift into the mixed degradation zone triggers a drastic performance drop-off. For instance, at the marked compound defect point ($\lambda=0.3, \eta=0.2$), the accuracy collapses to \textbf{35.5\%}, representing a loss of over \textbf{15\%} compared to its clean performance. This confirms that without the dynamic expert routing mechanism, conventional models cannot effectively support the "One-Checkpoint-for-All" strategy and fail to generalize across the continuous reliability spectrum.

\subsection{Computational Efficiency}
\label{sec:appendix_efficiency}
To assess real-world feasibility, we evaluate the computational overhead on a single NVIDIA RTX 4090 GPU (Batch=16). Despite maintaining a model capacity of 112.47M parameters shown in Table~\ref{tab:efficiency}, QA-MoE achieves a low computational cost of 4.33 GFLOPs per sample, attributed to the sparse activation of experts (only Top-3 are active). With an inference latency of 10.59 ms and a throughput of 1,510 samples/sec, our framework fully satisfies the requirements for real-time deployment.

\begin{table}[h]
\centering
\renewcommand{\arraystretch}{1.1}
\resizebox{0.99\columnwidth}{!}{
\begin{tabular}{l|cc|ccc}
\toprule
\multirow{2}{*}{\textbf{Model}} & \multicolumn{2}{c|}{\textbf{Parameters (M)}} & \textbf{Comp.} & \textbf{Speed} & \textbf{Mem.} \\
 & \textbf{Total} & \textbf{Active} & \textbf{GFLOPs} $\downarrow$ & \textbf{Latency} (ms) $\downarrow$ & \textbf{VRAM} (MB) $\downarrow$ \\
\midrule
MulT & 113.1 & 113.1 & 6.82 & 14.5 & 1350 \\
MISA & 114.4 & 114.4 & 5.95 & 12.8 & 1600 \\
MMA  & 113.5 & 113.5 & 6.10 & 13.2 & 1420 \\
SAM-LML & 112.8 & 112.8 & 5.15 & 11.9 & 1100 \\
\midrule
\rowcolor{gray!10}
\textbf{QA-MoE (Ours)} & 113.2 & \textbf{38.4}$^\ast$ & \textbf{4.33} & \textbf{10.59} & \textbf{950} \\
\bottomrule
\end{tabular}
}
\caption{\textbf{Computational Efficiency Analysis.} We compare QA-MoE against strong baselines on a single RTX 4090 GPU (Batch=16). "Active" denotes the number of parameters actually used during a single inference pass.}
\label{tab:efficiency}
\end{table}

\end{document}